\documentclass[a4paper]{article}%
\usepackage{amsmath}%
\usepackage{amsfonts}%
\usepackage{amssymb}%
\usepackage{graphicx}
\usepackage{fullpage}

\usepackage{math_defs}

\begin{document}

\title{Variational Bayes Factor Analysis for i-Vector Extraction}
\author{Jes\'{u}s Villalba\\\\
  Communications Technology Group (GTC),\\ Aragon Institute
  for Engineering Research (I3A),\\ University of Zaragoza, Spain\\
  \small \tt villalba@unizar.es}
\date{May 11, 2012}
\maketitle

\section{Introduction}

In this document we are going to derive the equations needed to
implement a Variational Bayes i-vector extractor. This can be used to
extract longer i-vectors reducing the risk of overfittig or to adapt
an i-vector extractor from a database to another with scarce
development data. This work is based on~\cite{Kenny2005} 
and~\cite{Bishop1999}.

\section{The Model}

\subsection{JFA}

\begin{figure}[th]
  \begin{center}
    \includegraphics[width=0.60\textwidth]
    {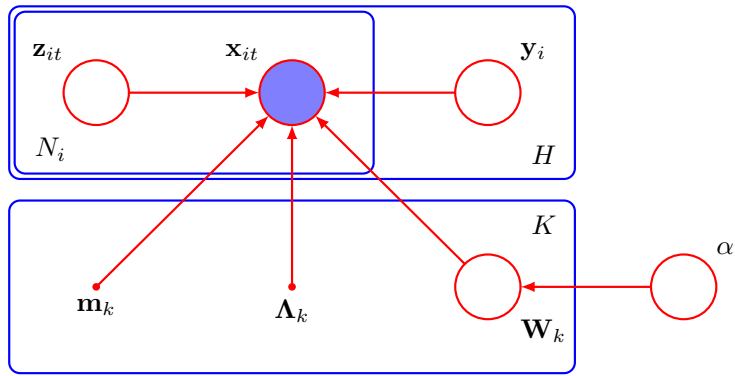}
  \end{center}
  \vspace{-0.5cm}
  \caption{BN for i-vector extractor.}
  \label{fig:bn_vbfa}
\end{figure}

Joint Factor Analysis for i-vector extraction is a linear generative
model represented in
Figure~\ref{fig:bn_vbfa}.

This model assumes that speech frames are generated by a special type
of mixture of factor analysers. 
An speech frame $\xvec_{it}$ of a session $i$
and generated by the component $k$ of the mixture model can be written as:
\begin{equation}
  \xvec_{it}=\mvec_k+\Wmat_k\yvec_i+\epsilon_{itk}
\end{equation}
where $\mvec_k$ is a session independent term, $\Wmat_k$ is a low-rank
factor loading
matrix, $\yvec_i$ is the factor vector, and $\epsilon_{itk}$ is a residual
term. 
The prior distribution for the variables:
\begin{align}
  \label{eq:splda_yprior}
  \yvec_i&\sim\Gauss{\yvec_i}{\zerovec}{\Imat} \\
  \label{eq:splda_eprior}
  \epsilon_{itk}&\sim\Gauss{\epsilon_{itk}}{\zerovec}{\iLambmat_k}
\end{align}
where $\mathcal{N}$ denotes a Gaussian distribution. 

This model differs from a standard mixture of FA in the way in
which the factors are tied. In traditional FA, we have a different
value of $\yvec$ for each frame and each component of the mixture of
the session. On the contrary, in this model we share the same value of
$\yvec$ for all the frames and mixture components of the same
session. 

We can define the session mean vector for component $k$ as
\begin{align}
  \Mvec_{ik}=\mvec_k+\Wmat_k\yvec_i \;.
\end{align}
In this manner, each frame is a session mean plus the residual term:
\begin{align}
  \xvec_{it}=\Mvec_{ik}+\epsilon_{itk} \;.
\end{align}

We find convening stacking the means and factor loading matrices of
all components to form a mean supervector:
\begin{align}
  \Mvec_{i}=\mvec+\Wmat\yvec_i
\end{align}

For this work, we are going to assume that $\mvec$ and $\Lambmat$ are
given. We estimate them by EM-iterations of simple GMM. Besides, we
assume that $\Prob{\zvec_{it}}$ are known and fixed. In practice, we
compute them using the GMM.

\subsection{Notation}

We define:
\begin{itemize}
\item Let $\Xmat_i$ be the frames of session $i$.
\item Let $\Xmat$ be the frames of all sessions.
\item Let $\Ymat$ be the factors of all sessions.
\item Let $d$ be the features dimension.
\item Let $n_y$ be the factor dimension.
\item Let $K$ be the number of components of the mixture of FA. 
\item Let $\Sigmat_k=\Lambmat_k^{-1}$.
\end{itemize}

\section{Sufficient statistics}

We define the statistics for segment $i$ and component $k$ as:
\begin{align}
  N_{ik}=\sum_{t}^{N_i}\Prob{\zvec_{itk}=1}
  \Fvec_{ik}=\sum_{t}^{N_i}\Prob{\zvec_{itk}=1}\xvec_{it}
\end{align} 

We define the normalized sufficient statistics for component $k$ as:
\begin{align}
  \Fbar_{ik}=&\sum_{t}^{N_i}\Prob{\zvec_{itk}=1}
  \Lambmat_k^{1/2}\left(\xvec_{it}-\mvec_k\right)
  =\Lambmat_k^{1/2}\left(\Fvec_{ik}-N_{ik}\mvec_k\right)
\end{align} 
If we normalize the sufficient statistics in mean and variance it is
the same as having a FA model with $\mvec=\zerovec$ and
$\Sigmat=\Imat$. As we assume that $\mvec$ and $\Sigmat$ are fixed,
doing that we can simplify the equations. 

We define, too:
\begin{align}
  \Nmat_i=
  \begin{bmatrix}
    N_{i1} \Imat_d  & \zerovec & \cdots & \zerovec \\
    \zerovec & N_{i2} \Imat_d  & \cdots & \zerovec \\
    \vdots  & \vdots & \ddots & \vdots \\
    \zerovec & \zerovec & \cdots & N_{iK} \Imat_d \\
  \end{bmatrix}
  , & \quad 
  \Fbar_i=
  \begin{bmatrix}
    \Fbar_{i1}\\
    \vdots\\
    \Fbar_{iK}
  \end{bmatrix}
\end{align}
where $\Imat_d$ is the identity
matrix of dimension $d$.

We define the global normalized statistic:
\begin{align}
  \Sbarmat_k=\sumiH\sumtNi \Prob{\zvec_{itk}=1}
  \mahP{\xvec_{it}}{\mvec_k}{\Lambmat_k}
\end{align}

\section{Conditional likelihood}

The likelihood of the data of session $i$ 
given the latent variables is
\begin{align}
  \lnProb{\Xmat_i|\Ymat,\Wmat,\mvec,\Lambmat}=&
  -\sumkK \frac{N_{ik}d}{2}\log(2\pi)
  -\med\trace\left(\sumkK\Sbarmat_{ik}\right)
  +\yvec_i^T\Wmat^T\Fbar_i-\med\yvec_i^T\Wmat^T\Nmat_i\Wmat\yvec_i
\end{align}

\section{Variational inference with Gaussian-Gamma priors}

\subsection{Model priors}

We introduce a \emph{hierarchical} prior $\Prob{\Wmat|\alphavec}$ over
the matrix $\Wmat$ governed by a $n_y$ dimensional vector of
hyperparameters where $n_y$ is the dimension of the factors. Each
hyperparameter controls one of the columns of the matrix $\Wmat$
through a conditional Gaussian distribution of the form:
\begin{align}
  \Prob{\Wmat|\alphavec}=
  \prod_{q=1}^{n_y}\left(\frac{\alpha_q}{2\pi}\right)^{Kd/2}
  \exp\left(-\med\alpha_q\wvec_q^T\wvec_q\right)
\end{align}
where $\wvec_q$ are the columns of $\Wmat$. Each $alpha_q$ controls
the inverse variance of the corresponding $\wvec_q$. If a particular
$\alpha_q$ has a posterior distribution concentrated at large
values, the corresponding $\wvec_q$ will tend to be small, and that
direction of the latent space will be effectively 'switched off'. 

We define a prior for the $\alphavec$:
\begin{align}
  \Prob{\alphavec}=\prod_{q=1}^{n_y}\Gammad{\alpha_q}{a}{b}
\end{align}
where $\mathcal{G}$ denotes the Gamma distribution. Bishop defines broad
priors setting $a=b=10^{-3}$.

\subsection{Variational distributions}
\label{sec:vbfa_vd}

We write the joint distribution of the latent variables:
\begin{align}
  \Prob{\Xmat,\Ymat,\Wmat,\alphavec|\mvec,\Lambmat,a,b}=
  \Prob{\Xmat|\Ymat,\Wmat,\mvec,\Lambmat}\Prob{\Ymat}
  \Prob{\Wmat|\alphavec}\Prob{\alphavec|a,b}
\end{align}
Following, the conditioning on $\left(\mvec,\Lambmat,a,b\right)$ will be dropped for
convenience. 

Now, we consider the partition of the posterior:
\begin{align}
  \Prob{\Ymat,\Wmat,\alphavec|\Xmat}\approx
  \q{\Ymat,\Wmat,\alphavec}=\q{\Ymat}\q{\Wmat}\q{\alphavec}
\end{align}

The optimum for $\qopt{\Ymat}$:
\begin{align}
  \lnqopt{\Ymat}=&
  \Expcond{\lnProb{\Xmat,\Ymat,\Wmat,\alphavec}}{\Wmat,\alphavec}+\const\\
  =&\Expcond{\lnProb{\Xmat|\Ymat,\Wmat}}{\Wmat}+\lnProb{\Ymat}+\const\\
  =&\sumiH \yvec_i^T\Exp{\Wmat}^T\Fbar_i
  -\med\yvec_i^T\left(\Imat+\sumkK N_{ik}\Exp{\Wmat_k^T\Wmat_k}\right)\yvec_i+\const
\end{align}
Therefore $\qopt{\Ymat}$ is a product of Gaussian distributions.
\begin{align}
  \qopt{\Ymat}=&\prodiH \Gauss{\yvec_i}{\ybarvec_i}{\iLmatyi}\\
  \Lmatyi=&\Imat+\sumkK N_{ik}\Exp{\Wmat_k^T\Wmat_k}\\
  \ybarvec_i=&\iLmatyi\Exp{\Wmat}^T\Fbar_i
\end{align}

The optimum for $\qopt{\Wmat}$:
\begin{align}
  \lnqopt{\Wmat}=&
  \Expcond{\lnProb{\Xmat,\Ymat,\Wmat,\alphavec}}{\Ymat,\alphavec}+\const\\
  =&\Expcond{\lnProb{\Xmat|\Ymat,\Wmat}}{\Ymat}
  +\Expcond{\lnProb{\Wmat|\alphavec}}{\alphavec}+\const\\
  =&\sumiH\left( \Exp{\yvec_i}^T\Wmat^T\Fbar_i
    -\med\Exp{\yvec_i^T\Wmat^T\Nmat_i\Wmat\yvec_i}\right)
  -\med \sum_{q=1}^{n_y} \Exp{\alpha_q}\wvec_q^T\wvec_q+\const\\
  =&\trace\left(\Wmat^T\Cmat
    -\med\sumkK\Wmat_k^T\Wmat_k\Rmat_k\right)
  -\med \sum_{q=1}^{n_y} \Exp{\alpha_q}\wvec_q^T\wvec_q+\const\\
  =&\sumkK \trace\left(\Wmat_k^T\Cmat_k
    -\med \Wmat_k^T\Wmat_k\Rmat_k\right)
  -\med \sumrd 
  \wvec_{kr}'^T\diag\left(\Exp{\alphavec}\right)\wvec_{kr}'
  +\const\\
  =&\sumkK\sumrd \trace\left(\wvec_{kr}'\Cmat_{kr}
    -\med \scatt{\wvec_{kr}'}\left(\Exp{\alphavec}+\Rmat_k\right)\right)
  +\const
\end{align}
where $\wvec_{kr}'$ is a column vector containing the $r^{th}$ row
of $\Wmat_k$,
\begin{align}
  \Wmat_k'=&\Wmat_k^T\\
  \Cmat=&\sumiH \Fbar_i\Exp{\yvec_i}^T\\
  \Rmat_k=&\sumiH N_{ik}\Exp{\scatt{\yvec_i}}
\end{align}
and $\Cmat_{kr}$ is the $r^th$ of the block of $\Cmat$ corresponding to
component $k$ (row $(k-1)*d+r$).

Then $\qopt{\Wmat}$ is a product of Gaussian distributions:
\begin{align}
  \qopt{\Wmat}=&\prodkK\prodrd
  \Gauss{\wvec_{kr}'}{\wbarvec_{kr}'}{\iLmatWk}\\
  \LmatWk=&\Exp{\alphavec}+\Rmat_k\\
  \wbarvec_{kr}'=&\iLmatWk\Cmat_{kr}^T
\end{align}

The optimum for $\qopt{\alphavec}$:
\begin{align}
  \lnqopt{\alphavec}=&
  \Expcond{\lnProb{\Xmat,\Ymat,\Wmat,\alphavec}}{\Ymat,\Wmat}+\const\\
  =&\Expcond{\lnProb{\Wmat|\alphavec}}{\Wmat}+\lnProb{\alphavec|a,b}+\const\\
  =& \sum_{q=1}^{n_y} \frac{Kd}{2}\ln \alpha_q -\med\alpha_q\Exp{\wvec_q^T\wvec_q}
  +(a-1)\ln \alpha_q-b\alpha_q+\const\\
  =& \sum_{q=1}^{n_y} \left(\frac{Kd}{2}+a-1\right)\ln \alpha_q 
  -\alpha_q\left(b+\med\Exp{\wvec_q^T\wvec_q}\right) +\const
\end{align}
Then
\begin{align}
  \label{eq:vbfa_apost}
  \qopt{\alphavec}=&\prod_{q=1}^{n_y}\Gammad{\alpha_q}{a^\prime}{b_q^\prime}\\
  a^\prime=&a+\frac{Kd}{2}\\
  b_q^\prime=&b+\med\Exp{\wvec_q^T\wvec_q}
\end{align}

We evaluate the expectations:
\begin{align}
  \Exp{\alpha_q}=&\frac{a^\prime}{b_q^\prime}\\
  \Exp{\Wmat}=&
  \begin{bmatrix}
    \wbarvec_{11}'^T\\
    \wbarvec_{12}'^T\\
    \vdots\\
    \wbarvec_{Kd}'^T
  \end{bmatrix}\\
  \Exp{\wvec_q^T\wvec_q}=&\sumkK\sumrd\Exp{\wvec_{krq}'^T\wvec_{krq}'}\\
  =&\sumkK\sumrd\iLmat_{\Wmat_k qq}+\wbarvec_{rkq}'^2\\
  =&\sumkK d \iLmat_{\Wmat_k qq}+\sumrd\wbarvec_{rkq}'^2\\
  \Exp{\Wmat_k^T\Wmat_k}=&\Exp{\Wmat_k'\Wmat_k'^T}\\
  =&d\iLmatWk+\Exp{\Wmat_k}^T\Exp{\Wmat_k}
\end{align}

\subsection{Variational lower bound}
\label{sec:vbfa_lb}

The lower bound is given by
\begin{align}
  \lowb=&\Expcond{\lnProb{\Xmat|\Ymat,\Wmat}}{\Ymat,\Wmat}
  +\Expcond{\lnProb{\Ymat}}{\Ymat}
  +\Expcond{\lnProb{\Wmat|\alphavec}}{\Wmat,\alphavec}
  +\Expcond{\lnProb{\alphavec}}{\alphavec}\nonumber\\
  &-\Expcond{\lnq{\Ymat}}{\Ymat}
  -\Expcond{\lnq{\Wmat}}{\Wmat}
  -\Expcond{\lnq{\alphavec}}{\alphavec}
\end{align}

The term $\Expcond{\lnProb{\Xmat|\Ymat,\Wmat}}{\Ymat,\Wmat}$:
\begin{align}
  \Expcond{\lnProb{\Xmat|\Ymat,\Wmat}}{\Ymat,\Wmat}=&
  -\sumkK \frac{N_{k}d}{2}\log(2\pi)
  -\med\trace\left(\sumkK\Sbarmat_k\right)\nonumber\\
  &+\sumiH\Exp{\yvec_i}^T\Exp{\Wmat}^T\Fbar_i
  -\med\sumkK\sumiH\trace\left(N_{ik}\Exp{\Wmat_k^T\Wmat_k}\Exp{\scatt{\yvec_i}}\right)\\
  =&-\sumkK \frac{N_{k}d}{2}\log(2\pi)
  -\med\trace\left(\sumkK\Sbarmat_k\right)\nonumber\\
  &-\med\trace\left(-2\Exp{\Wmat}^T\Cmat+\sumkK\Exp{\Wmat_k^T\Wmat_k}\Rmat_k\right)
\end{align}

The term $\Expcond{\lnProb{\Ymat}}{\Ymat}$:
\begin{align}
  \Expcond{\lnProb{\Ymat}}{\Ymat}=&
  -\frac{H n_y}{2}\ln(2\pi)-\med\trace\left(\sumiH
    \Exp{\scatt{\yvec_i}}\right)\\
  =&-\frac{H n_y}{2}\ln(2\pi)-\med\trace\left(\Rhomat\right)
\end{align}
where
\begin{align}
  \Rhomat=\sumiH \Exp{\scatt{\yvec_i}}
\end{align}

The term $\Expcond{\lnProb{\Wmat|\alphavec}}{\Wmat,\alphavec}$:
\begin{align}
  \Expcond{\lnProb{\Wmat|\alphavec}}{\Wmat,\alphavec}=&
  -\frac{n_y Kd}{2}\ln(2\pi)+\frac{Kd}{2}\sum_{q=1}^{n_y}\Exp{\ln\alpha_q}
  -\med\sum_{q=1}^{n_y} \Exp{\alpha_q}\Exp{\wvec_q^T\wvec_q}
\end{align}
where
\begin{align}
  \Exp{\ln\alpha_q}=\psi(a^\prime)-\ln b_q^\prime
\end{align}
where $\psi$ is the digamma function.

The term $\Expcond{\lnProb{\alphavec}}{\alphavec}$:
\begin{align}
  \Expcond{\lnProb{\alphavec}}{\alphavec}=&
  n_y \left(a\ln b-\ln\gammaf{a}\right)+
  \sum_{q=1}^{n_y}(a-1)\Exp{\ln\alpha_q}-b\Exp{\alpha_q}\\
  =& n_y \left(a\ln b-\ln\gammaf{a}\right)+
  (a-1)\sum_{q=1}^{n_y}\Exp{\ln\alpha_q}-b\sum_{q=1}^{n_y}\Exp{\alpha_q}
\end{align}

The term $\Expcond{\lnq{\Ymat}}{\Ymat}$:
\begin{align}
  \Expcond{\lnq{\Ymat}}{\Ymat}=&
  -\frac{Hn_y}{2}\ln(2\pi)+\med\sumiH\lndet{\Lmatyi}
  -\med\trace\left(\Lmatyi\Exp{\scattp{\yvec_i-\ybarvec_i}}\right)\\
  =&-\frac{Hn_y}{2}\ln(2\pi)+\med\sumiH\lndet{\Lmatyi} \nonumber\\
  &-\med\sumiH\trace\left(\Lmatyi\left(
      \Exp{\scatt{\yvec_i}}
      -\ybarvec_i\Exp{\yvec_i}^T-\Exp{\yvec_i}\ybarvec_i^T
      +\scatt{\ybarvec_i}\right)\right)\\
  =&-\frac{Hn_y}{2}\ln(2\pi)+\med\sumiH\lndet{\Lmatyi}
  -\med\sumiH\trace\left(\Imat\right)\\
  =&-\frac{Hn_y}{2}(\ln(2\pi)+1)+\med\sumiH\lndet{\Lmatyi}
\end{align}

The term $\Expcond{\lnq{\Wmat}}{\Wmat}$:
\begin{align}
  \Expcond{\lnq{\Wmat}}{\Wmat}=&
  -\frac{Kdn_y}{2}\ln(2\pi)+\frac{d}{2}\sumkK\lndet{\LmatWk}\nonumber\\
  &-\med\sumkK\sumrd
  \trace\left(\LmatWk\Exp{\scattp{\wvec_{kr}'-\wbarvec_{kr}'}}\right)\\
  =&-\frac{Kdn_y}{2}\left(\ln(2\pi)+1\right)+\frac{d}{2}\sumkK\lndet{\LmatWk}
\end{align}

The term $\Expcond{\lnq{\alphavec}}{\alphavec}$:
\begin{align}
  \Expcond{\lnq{\alphavec}}{\alphavec}=&
  -\sum_{q=1}^{n_y}\Entrop{\q{\alpha_q}}\\
  =&\sum_{q=1}^{n_y}(a^\prime-1)\psi(a^\prime)
  +\ln b_q^\prime-a^\prime-\ln\gammaf{a^\prime}\\
  =&n_y\left((a^\prime-1)\psi(a^\prime)-a^\prime-\ln\gammaf{a^\prime}\right)
  +\sum_{q=1}^{n_y}\ln b_q^\prime
\end{align}

\subsection{Hyperparameter optimization}

We can set the Hyperparameters manually or estimate them from the
development data maximizing the lower bound.

We derive for $a$:
\begin{align}
  \frac{\partial\lowb}{\partial a}=&
  n_y\left(\ln b-\psi(a)\right)
  +\sum_{q=1}^{n_y}\Exp{\ln\alpha_q}=0 \quad \implies\\
  \psi(a)=&\ln b+\frac{1}{n_y}\sum_{q=1}^{n_y}\Exp{\ln\alpha_q}
\end{align}

We derive for $b$:
\begin{align}
  \frac{\partial\lowb}{\partial b}=&
  \frac{n_y a}{b}-\sum_{q=1}^{n_y}\Exp{\alpha_q}=\zerovec\quad \implies\\
  b=&\left( \frac{1}{n_y a}\sum_{q=1}^{n_y}\Exp{\alpha_q}\right)^{-1}
\end{align}

We solve these equation with the procedure described
in~\cite{Beal2003}. We write
\begin{align}
  \psi(a)=&\ln b+c\\
  b=&\frac{a}{d}
\end{align}
where
\begin{align}
  c=&\frac{1}{n_y}\sum_{q=1}^{n_y}\Exp{\ln\alpha_q}\\
  d=&\frac{1}{n_y}\sum_{q=1}^{n_y}\Exp{\alpha_q}
\end{align}
Then
\begin{align}
  f(a)=\psi(a)-\ln a + \ln d -c=0
\end{align}

We can solve for $a$ using Newton-Rhaphson iterations:
\begin{align}
  a_{new}=&a-\frac{f(a)}{f^\prime(a)}=\\
  =&a\left(1-\frac{\psi(a)-\ln a + \ln d -c}{a\psi^\prime(a)-1}\right)
\end{align}
This algorithm does not assure that $a$ remains positive. We can put a
minimum value for $a$. Alternatively we can solve the equation for
$\tilde{a}$ such as $a=exp(\tilde{a})$.
\begin{align}
  \tilde{a}_{new}=&\tilde{a}-\frac{f(\tilde{a})}{f^\prime(\tilde{a})}=\\
  =&\tilde{a}-\frac{\psi(a)-\ln a + \ln d -c}{\psi^\prime(a)a-1}
\end{align}
Taking exponential in both sides:
\begin{align}
  a_{new}=a\exp\left(-\frac{\psi(a)-\ln a + \ln d -c}{\psi^\prime(a)a-1}\right)
\end{align}

\subsection{Minimum divergence}

We assume a more general prior for the hidden variables:
\begin{align}
  \Prob{\yvec}=\Gauss{\yvec}{\muvecy}{\iLambmaty}
\end{align}
To minimize the divergence we maximize the part of $\lowb$ that depends
on $\muvecy$:
\begin{align}
  \lowb(\muvecy,\Lambmaty)=&\sumiH \ExpcondY{\ln
    \Gauss{\yvec}{\muvecy}{\iLambmaty}}
\end{align}

The, we get
\begin{align}
\muvecy=&\frac{1}{H}\sumiM\ExpcondY{\yvec_i}\\
\Sigmaty=&\Lambmaty^{-1}
=\frac{1}{H}\sumiH\ExpcondY{\scattp{\yvec_i-\muvecy}}\\
=&\frac{1}{H}\sumiH\ExpcondY{\scatt{\yvec_i}}-\scatt{\muvecy}
\end{align}

We have a transform $\yvec=\phi(\yvec^\prime)$ such as
$\yvec^\prime$ has a standard prior:
\begin{align}
  \yvec=&\muvecy+(\Sigmaty^{1/2})^{T}\yvec^\prime
\end{align}

Now, we get $\q{\Wmat}$ such us if we apply the transform
$\yvec'=\phivec^{-1}(\yvec)$, the term 
$\Exp{\lnProb{\Xmat|\Ymat,\Wmat}}$ of 
$\lowb$ remains constant:
\begin{align}
  \wbarvec_{kr}'\leftarrow&\Sigmaty^{1/2}\wbarvec_{kr}'\\
  \iLmatWk\leftarrow&\Sigmaty^{1/2}\iLmatWk(\Sigmaty^{1/2})^{T}\\
  \LmatWk\leftarrow&
  \left((\Sigmaty^{1/2})^{-1}\right)^T\LmatWk(\Sigmaty^{1/2})^{-1}
\end{align}

\section{Variational inference with full covariance priors}

\subsection{Model priors}

Lets assume that we compute
the posterior of $\Wmat$ given a development database with a large
amount of data. If we want to compute the posterior $\Wmat$ for a
small database we could use the posterior given the large database as
prior. Thus, we take a prior distribution for $\Wmat$
\begin{align}
  \Prob{\Wmat}=\prodkK\prodrd
  \Gauss{\wvec_{kr}'}{\wbarvec_{0 kr}'}{\iLmatWdk}
\end{align}
where $\wbarvec_{0 kr}'$, $\iLmatWdk$ are parameters computed with
the large dataset.

\subsection{Variational distributions}

The joint distribution of the latent variables:
\begin{align}
  \Prob{\Xmat,\Ymat,\Wmat}=
  \Prob{\Xmat|\Ymat,\Wmat,\mvec,\Lambmat}\Prob{\Ymat}
  \Prob{\Wmat}
\end{align}

We approximate the posterior by:
\begin{align}
  \Prob{\Ymat,\Wmat|\Xmat}\approx
  \q{\Ymat,\Wmat}=\q{\Ymat}\q{\Wmat}
\end{align}

The optimum for $\qopt{\Ymat}$ is the same as in section~\ref{sec:vbfa_vd}.

The optimum for $\qopt{\Wmat}$ is
\begin{align}
  \lnqopt{\Wmat}=&
  \Expcond{\lnProb{\Xmat,\Ymat,\Wmat}}{\Ymat}+\const\\
  =&\Expcond{\lnProb{\Xmat|\Ymat,\Wmat}}{\Ymat}
  +\lnProb{\Wmat}+\const\\
  =&\sumkK\sumrd \trace\left(\wvec_{kr}'\Cmat_{kr}
    -\med \scatt{\wvec_{kr}'}\Rmat_k \right)
  -\med \mahP{\wvec_{kr}'}{\wbarvec_{0 kr}'}{\LmatWdk}
  +\const\\
  =&\sumkK\sumrd \trace\left(
    \wvec_{kr}'\left(\wbarvec_{0 kr}'^T\LmatWdk+\Cmat_{kr}\right)
    -\med \scatt{\wvec_{kr}'}\left(\LmatWdk+\Rmat_k\right)\right)
  +\const\\
\end{align}

Therefore, the $\qopt{\Wmat}$ is, again, a product of Gaussian distributions:
\begin{align}
  \qopt{\Wmat}=&\prodkK\prodrd
  \Gauss{\wvec_{kr}'}{\wbarvec_{kr}'}{\iLmatWk}\\
  \LmatWk=&\LmatWdk+\Rmat_k\\
  \wbarvec_{kr}'=&\iLmatWk\left(\LmatWdk\wbarvec_{0 kr}'+\Cmat_{kr}^T\right)
\end{align}

\subsection{Variational lower bound}

The lower bound is given by
\begin{align}
  \lowb=&\Expcond{\lnProb{\Xmat|\Ymat,\Wmat}}{\Ymat,\Wmat}
  +\Expcond{\lnProb{\Ymat}}{\Ymat}
  +\Expcond{\lnProb{\Wmat}}{\Wmat}
  -\Expcond{\lnq{\Ymat}}{\Ymat}
  -\Expcond{\lnq{\Wmat}}{\Wmat}
\end{align}

The term $\Expcond{\lnProb{\Wmat}}{\Wmat}$:
\begin{align}
  \Expcond{\lnProb{\Wmat}}{\Wmat}=&
  -\frac{n_y Kd}{2}\ln(2\pi)+\frac{d}{2}\sumkK\lndet{\LmatWdk} \nonumber\\
  &-\med\sumkK\sumrd\trace\left(\LmatWdk
    \Exp{\scattp{\wvec_{kr}'-\wbarvec_{0 kr}'}}\right)\\
  =&-\frac{n_y Kd}{2}\ln(2\pi)+\frac{d}{2}\sumkK\lndet{\LmatWdk} \nonumber\\
  &-\med\sumkK \sumrd \trace\left(\LmatWdk\left(\iLmatWk
      +\scatt{\wbarvec_{kr}'}
      -\wbarvec_{0 kr}'\wbarvec_{kr}'^T
      -\wbarvec_{kr}'\wbarvec_{0 kr}'^T
      +\scatt{\wbarvec_{0 kr}'}\right)\right)\\
  =&-\frac{n_y Kd}{2}\ln(2\pi)+\frac{d}{2}\sumkK\lndet{\LmatWdk} \nonumber\\
  &-\frac{d}{2}\sumkK  \trace\left(\LmatWdk\iLmatWk\right)
  -\med\sumkK \sumrd 
  \mahP{\wbarvec_{kr}'}{\wbarvec_{0 kr}'}{\LmatWdk}\\
  =&-\frac{n_y Kd}{2}\ln(2\pi)+\frac{d}{2}\sumkK\lndet{\LmatWdk} \nonumber\\
  &-\frac{d}{2}\sumkK  \trace\left(\LmatWdk\iLmatWk\right)
  -\med\sumkK \trace\left(\LmatWdk  
    \sumrd\scattp{\wbarvec_{kr}'-\wbarvec_{0 kr}'}\right)
\end{align}

The rest of terms are the same as the ones in section~\ref{sec:vbfa_lb}.

\section{Variational inference  with
  Gaussian-Gamma priors for high rank $\Wmat$ }

The amount of memory needed for the factor analyser grows quadratically
with the dimension of the factor vector $n_y$. Due to that, we are
limited to use small i-vectors ($n_y<1000$). We are going to modify
the variational partition function so that the memory grows linearly
with the number of factors. We derive the equations for the case of
Gaussian-Gamma prior for $\Wmat$.

\subsection{Variational distributions}

We choose the partition function:
\begin{align}
  \Prob{\Ymat,\Wmat,\alphavec|\Xmat}\approx
  \q{\Ymat,\Wmat,\alphavec}=\prodpP\q{\Ymatsupp}\q{\Wmatsupp}\q{\alphavec}
\end{align}
where
\begin{align}
  \Ymat=
  \begin{bmatrix}
    \Ymat^{(1)}\\
    \Ymat^{(2)}\\
    \vdots\\
    \Ymat^{(P)}
  \end{bmatrix}
\end{align}
We define the blocks $\Wmat_k$, $\Wmat^{(p)}$ and $\Wmat_k^{(p)}$ of $\Wmat$ such as
\begin{align}
  \Wmat=
  \begin{bmatrix}
    \Wmat_1 \\
    \Wmat_2 \\
    \vdots \\
    \Wmat_K
  \end{bmatrix}
  =
  \begin{bmatrix}
    \Wmat^{(1)} & \Wmat^{(2)} & \hdots & \Wmat^{(P)}
  \end{bmatrix}
  =
  \begin{bmatrix}
    \Wmat_1^{(1)} & \Wmat_1^{(2)} & \hdots & \Wmat_1^{(P)} \\
    \Wmat_2^{(1)} & \Wmat_2^{(2)} & \hdots & \Wmat_2^{(P)} \\ 
    \vdots & \vdots & \ddots & \vdots \\ 
    \Wmat_K^{(1)} & \Wmat_K^{(2)} & \hdots & \Wmat_K^{(P)} \\
  \end{bmatrix}
\end{align}

This partition function assumes that there are groups of components of
the i-vectors that are independent between them in the posterior. For
example, the components in $\Ymat_1$ would be independent from the
components in $\Ymat_2$ but the components inside $\Ymat_1$ would be
dependent between them. We are going to assume that every group has
the same number of components $\tilde{n}_y=n_y/P$.

The optimum for $\qopt{\Ymatsupp}$:
\begin{align}
  \lnqopt{\Ymatsupp}=&
  \Expcond{\lnProb{\Xmat,\Ymat,\Wmat,\alphavec}}
  {\Wmat,\alphavec,\Ymatsup{s\neq p}}+\const\\
  =&\Expcond{\lnProb{\Xmat|\Ymat,\Wmat}}{\Wmat,\Ymatsup{s\neq p}}
  +\lnProb{\Ymatsupp}+\const\\
  =&\sumiH \Expcond{\yvec_i}{\Ymatsup{s\neq p}}^T\Exp{\Wmat}^T\Fbar_i
  \nonumber\\
  &-\med \sumkK N_{ik} \Expcond{ \yvec_i^T\Wmat_k^T\Wmat_k\yvec_i}
  {\Wmat,\Ymatsup{s\neq p}}-\med\yvecsuppT\yvecsupp+\const\\
  =&\sumiH \yvecsuppT_i\Exp{\Wmatsupp}^T\Fbar_i
  \nonumber\\
  &-\med \sumkK N_{ik} 
  \Expcond{\sumnP\summP
    \yvecsupT{n}_i\WmatsupT{n}_k\Wmatsup{m}_k\yvecsup{m}_i}
  {\Wmat,\Ymatsup{s\neq p}}-\med\yvecsuppT\yvecsupp+\const\\
  =&\sumiH \yvecsuppT_i\Exp{\Wmatsupp}^T\Fbar_i
  -\med \yvecsuppT_i 
  \left(\Imat+\sumkK N_{ik} \Exp{\WmatsuppT_k\Wmatsupp_k} \right)
  \yvecsupp_i
  \nonumber\\
  &- \yvecsuppT_i \sumkK N_{ik} \sum_{n \neq p} \Exp{\WmatsuppT_k\Wmatsup{n}_k}
  \Exp{\yvecsup{n}_i}
  +\const\\
  =&\sumiH \yvecsuppT_i \sumkK 
  \left(\Exp{\Wmatsupp_k}^T\Fbar_{ik}
    - N_{ik} \sum_{n \neq p} \Exp{\WmatsuppT_k\Wmatsup{n}_k}
    \Exp{\yvecsup{n}_i}\right)
  \nonumber\\
  &-\med \yvecsuppT_i 
  \left(\Imat+\sumkK N_{ik} \Exp{\WmatsuppT_k\Wmatsupp_k} \right)
  \yvecsupp_i
  +\const\\
  =&\sumiH \yvecsuppT_i \sumkK 
  \left(\Exp{\Wmatsupp_k}^T\Fbar_{ik}
    - N_{ik} \sum_{n \neq p} \Exp{\Wmatsupp_k}^T\Exp{\Wmatsup{n}_k}
    \Exp{\yvecsup{n}_i}\right)
  \nonumber\\
  &-\med \yvecsuppT_i 
  \left(\Imat+\sumkK N_{ik} \Exp{\WmatsuppT_k\Wmatsupp_k} \right)
  \yvecsupp_i
  +\const\\
  =&\sumiH \yvecsuppT_i 
  \Exp{\Wmatsupp}^T\left(\Fbar_{i}
    - \Nmat_i \sum_{n \neq p} \Exp{\Wmatsup{n}}
    \Exp{\yvecsup{n}_i}\right)
  \nonumber\\
  &-\med \yvecsuppT_i 
  \left(\Imat+\sumkK N_{ik} \Exp{\WmatsuppT_k\Wmatsupp_k} \right)
  \yvecsupp_i
  +\const
\end{align}
Therefore $\qopt{\Ymatsupp}$ is a product of Gaussian distributions.
\begin{align}
  \qopt{\Ymatsupp}=&\prodiH \Gauss{\yvecsupp_i}{\ybarvecsupp_i}{\iLmatsuppyi}\\
  \Lmatsuppyi=&\Imat+\sumkK N_{ik} \Exp{\WmatsuppT_k\Wmatsupp_k}\\
  \ybarvecsupp_i=&\iLmatsuppyi \Fbar_{i}
  - \Nmat_i \sum_{n \neq p} \Exp{\Wmatsup{n}}
  \Exp{\yvecsup{n}_i}
\end{align}

The optimum for $\qopt{\Wmatsupp}$:
\begin{align}
  \lnqopt{\Wmatsupp}=&
  \Expcond{\lnProb{\Xmat,\Ymat,\Wmat,\alphavec}}
  {\Ymat,\Wmatsup{s\neq p},\alphavec}+\const\\
  =&\Expcond{\lnProb{\Xmat|\Ymat,\Wmat}}{\Ymat,\Wmatsup{s\neq p}}
  +\Expcond{\lnProb{\Wmat|\alphavec}}{\Wmatsup{s\neq p},\alphavec}+\const\\
  =&\sumiH\left( \Exp{\yvecsupp_i}^T\WmatsuppT\Fbar_i
    -\med\Expcond{\yvec_i^T\Wmat^T\Nmat_i\Wmat\yvec_i}
    {\Ymat,\Wmatsup{s\neq p}}\right) \nonumber\\
  &-\med \sum_{q=1}^{\tilde{n}_y} \Exp{\alphasupp_q}\wvecsuppT_q\wvecsupp_q+\const\\
  =&\sumiH\left( \Exp{\yvecsupp_i}^T\WmatsuppT\Fbar_i
    -\med\sumkK N_{ik} 
    \Expcond{\sumnP\summP
      \yvecsupT{n}_i\WmatsupT{n}_k\Wmatsup{m}_k\yvecsup{m}_i}
    {\Ymat,\Wmatsup{s\neq p}}\right) \nonumber\\
  &-\med \sum_{q=1}^{\tilde{n}_y} \Exp{\alphasupp_q}\wvecsuppT_q\wvecsupp_q+\const\\
  =&\trace\left(\WmatsuppT\sumiH\Fbar_i\Exp{\yvecsupp_i}^T
    -\med \sumkK\sumnP\summP 
    \Expcond{\WmatsupT{n}_k\Wmatsup{m}}{\Wmatsup{s\neq p}}
    \sumiH N_{ik} \Exp{\yvecsup{m}_i\yvecsupT{n}_i}\right) \nonumber\\
  &-\med \sum_{q=1}^{\tilde{n}_y} \Exp{\alphasupp_q}\wvecsuppT_q\wvecsupp_q+\const\\
  =&\trace\left(\WmatsuppT\sumiH\left(
      \Fbar_i-\Nmat_i\sum_{n\neq p}\Exp{\Wmatsup{n}}\Exp{\yvecsup{n}_i}\right)
    \Exp{\yvecsupp_i}^T \right. \nonumber\\ 
  &\left.-\med \sumkK \WmatsuppT_k\Wmatsupp
    \sumiH N_{ik} \Exp{\yvecsupp_i\yvecsuppT_i}\right) \nonumber\\
  &-\med \sum_{q=1}^{\tilde{n}_y} \Exp{\alphasupp_q}\wvecsuppT_q\wvecsupp_q+\const\\
  =&\trace\left(\WmatsuppT\Cmatsupp
    -\med\sumkK\WmatsuppT_k\Wmatsupp_k\Rmatsupp_k\right) 
  -\med \sum_{q=1}^{\tilde{n}_y} \Exp{\alphasupp_q}\wvecsuppT_q\wvecsupp_q+\const\\
  =&\sumkK \trace\left(\WmatsuppT\Cmatsupp
    -\med\WmatsuppT_k\Wmatsupp_k\Rmatsupp_k\right) \nonumber\\
  &-\med \sumrd 
  \wvecpsuppT_{kr}\diag\left(\Exp{\alphavecsupp}\right)\wvecpsupp_{kr}
  +\const\\
  =&\sumkK\sumrd \trace\left(\wvecpsupp_{kr}\Cmatsupp_{kr}
    -\med \wvecpsupp_{kr}\wvecpsuppT_{kr}
    \left(\Exp{\alphavecsupp}+\Rmatsupp_k\right)\right)
  +\const
\end{align}
where $\wvecpsupp_{kr}$ is a column vector containing the $r^{th}$ row
of $\Wmatsupp_k$,
\begin{align}
  \Cmatsupp=&\sumiH\left(
    \Fbar_i-\Nmat_i\sum_{n\neq p}\Exp{\Wmatsup{n}}\Exp{\yvecsup{n}_i}\right)
  \Exp{\yvecsupp_i}^T \\
  \Rmatsupp_k=&\sumiH N_{ik} \Exp{\yvecsupp_i\yvecsuppT_i}
\end{align}
and $\Cmatsupp_{kr}$ is the $r^th$ of the block of $\Cmatsupp$ corresponding to
component $k$ (row $(k-1)*d+r$).

Then $\qopt{\Wmatsupp}$ is a product of Gaussian distributions:
\begin{align}
  \qopt{\Wmatsupp}=&\prodkK\prodrd
  \Gauss{\wvecpsupp_{kr}}{\wbarvecpsupp_{kr}}{\iLmatsuppWk}\\
  \LmatsuppWk=&\Exp{\alphavecsupp}+\Rmatsupp_k\\
  \wbarvecpsupp_{kr}=&\iLmatsuppWk\Cmat^{(p)T}_{kr}
\end{align}

The optimum for $\qopt{\alphavec}$ is the same as in
equation~\eqref{eq:vbfa_apost}. 

We need to evaluate the expectations:
\begin{align}
  \Exp{\wvecsuppT_q\wvecsupp_q}=&
  \sumkK d \Lmat_{\Wmat_k qq}^{(p)^{-1}}+\sumrd\wbarvec_{rkq}'^{(p)^2}\\
  \Exp{\WmatsuppT_k\WmatsuppT_k}=&
  \iLmatsuppWk+\Exp{\Wmatsupp_k}^T\Exp{\Wmatsupp_k}
\end{align}

\subsection{Variational lower bound}

The lower bound is given by
\begin{align}
  \lowb=&\Expcond{\lnProb{\Xmat|\Ymat,\Wmat}}{\Ymat,\Wmat}
  +\Expcond{\lnProb{\Ymat}}{\Ymat}
  +\Expcond{\lnProb{\Wmat|\alphavec}}{\Wmat,\alphavec}
  +\Expcond{\lnProb{\alphavec}}{\alphavec}\nonumber\\
  &-\sumpP\Expcond{\lnq{\Ymatsupp}}{\Ymatsupp}
  -\sumpP\Expcond{\lnq{\Wmatsupp}}{\Wmatsupp}
  -\Expcond{\lnq{\alphavec}}{\alphavec}
\end{align}

The term $\Expcond{\lnProb{\Xmat|\Ymat,\Wmat}}{\Ymat,\Wmat}$:
\begin{align}
  \Expcond{\lnProb{\Xmat|\Ymat,\Wmat}}{\Ymat,\Wmat}=&
  -\sumkK \frac{N_{k}d}{2}\log(2\pi)
  -\med\trace\left(\sumkK\Sbarmat_k\right)
  +\sumiH\Exp{\yvec_i}^T\Exp{\Wmat}^T\Fbar_i \nonumber\\
  &-\med\sumkK\sumiH \sumnP \summP \trace\left(
    N_{ik}\Exp{\WmatsupT{n}_k\Wmatsup{m}_k}
    \Exp{\yvecsup{m}_i\yvecsupT{n}_i}\right)\\
  =&-\sumkK \frac{N_{k}d}{2}\log(2\pi)
  -\med\trace\left(\sumkK\Sbarmat_k\right)
  +\trace\left(\Exp{\Wmat}^T\Cmat\right)\nonumber\\
  &-\med\sumkK\sumnP
  \trace\left(\Exp{\WmatsupT{n}_k\Wmatsup{n}_k}\Rmatsup{n}_k \right.
  \nonumber\\
  &\left.+2\sum_{m=n+1}^P
    \Exp{\Wmatsup{n}_k}^T\Exp{\Wmatsup{m}_k}\Rmatsup{m,n}_k \right)
\end{align}
where
\begin{align}
  \Rmatsup{m,n}_k=\sumiH N_{ik} \Exp{\yvecsup{m}_i}\Exp{\yvecsup{n}_i}^T
\end{align}

The term $\Expcond{\lnProb{\Ymat}}{\Ymat}$:
\begin{align}
  \Expcond{\lnProb{\Ymat}}{\Ymat}=&
  -\frac{H n_y}{2}\ln(2\pi)-\med\trace\left(\sumiH
    \Exp{\scatt{\yvec_i}}\right)\\
  =&-\frac{H n_y}{2}\ln(2\pi)-\med\sumpP\trace\left(\sumiH
    \Exp{\yvecsupp_i\yvecsuppT_i}\right)\\
  =&-\frac{H n_y}{2}\ln(2\pi)-\med\sumpP\trace\left(\Rhomatsupp\right)
\end{align}
where
\begin{align}
  \Rhomatsupp=\sumiH \Exp{\yvecsupp_i\yvecsuppT_i}
\end{align}

The term $\Expcond{\lnq{\Ymatsupp}}{\Ymatsupp}$:
\begin{align}
  \Expcond{\lnq{\Ymatsupp}}{\Ymatsupp}=&
  -\frac{H\tilde{n}_y}{2}(\ln(2\pi)+1)+\med\sumiH\lndet{\Lmatsuppyi}
\end{align}

The term $\Expcond{\lnq{\Wmatsupp}}{\Wmatsupp}$:
\begin{align}
  \Expcond{\lnq{\Wmatsupp}}{\Wmatsupp}=&
  -\frac{Kd\tilde{n}_y}{2}\left(\ln(2\pi)+1\right)+\frac{d}{2}\sumkK\lndet{\LmatsuppWk}
\end{align}

The rest of terms are the same as in section~\ref{sec:vbfa_lb}.

\bibliographystyle{IEEEbib}
\bibliography{villalba}

\end{document}